\documentclass{article}
\usepackage{arxiv}
\usepackage{graphicx}
\usepackage{subcaption}
\usepackage{amsmath,amssymb}
\usepackage{mathptmx}
\usepackage[group-separator={,}]{siunitx}
\usepackage{multirow}
\usepackage{booktabs}

\newcolumntype{L}[1]{>{\raggedright\let\newline\\\arraybackslash\hspace{0pt}}p{#1}}
\newcolumntype{C}[1]{>{\centering\let\newline\\\arraybackslash\hspace{0pt}}p{#1}}
\newcolumntype{R}[1]{>{\raggedleft\let\newline\\\arraybackslash\hspace{0pt}}p{#1}}

\title{RED-NET: A Recursive Encoder-Decoder Network for Edge Detection}
\author{
  Truc Le   Yuyan Li    Ye Duan \\
  Department of Computer Science\\
  University of Missouri\\
  Columbia, MO 65211 \\
  \texttt{http://engineers.missouri.edu/duanye/} \\
}

\begin{document}
\maketitle

\begin{abstract}
In this paper, we introduce RED-NET: A Recursive Encoder-Decoder Network with Skip-Connections for edge detection in natural images. The proposed network is a novel integration of a Recursive Neural Network with an Encoder-Decoder architecture. The recursive network enables us to increase the network depth without increasing the number of parameters. Adding skip-connections between encoder and decoder helps the gradients reach all the layers of a network more easily and allows information related to finer details in the early stage of the encoder to be fully utilized in the decoder. Based on our extensive experiments on popular boundary detection datasets including BSDS500 \cite{Arbelaez2011}, NYUD \cite{Silberman2012} and Pascal Context \cite{Mottaghi2014}, RED-NET significantly advances the  state-of-the-art on edge detection regarding standard evaluation metrics such as Optimal Dataset Scale (ODS) F-measure, Optimal Image Scale (OIS) F-measure, and Average Precision (AP).
\end{abstract}

\section{Introduction}
\label{sec:introduction}
Edge detection has been a cornerstone and long-standing problem in computer vision since the early 1970's \cite{Chien1974,Fram1975,Robinson1977} and is essential  for a variety of tasks such as object recognition \cite{Ullman1991,Ferrari2008}, segmentation \cite{Malik2001,Arbelaez2011,Kokkinos2016,Kass1988}, etc. Initially considered as a low-level task, researchers now generally agree hat high-level visual context such as the perception of objects play an important role in edge detection \cite{Arbelaez2011}.

Inspired by the success of deep convolutional neural networks (DCNN) in computer vision problems such as image classification \cite{Krizhevsky2012,Simonyan2014,Szegedy2015}, object detection \cite{Ren2016}, image segmentation \cite{Farabet2013,Sharma2015,Chen2015,Long2015,Kendall2015,Noh2015}, normal estimation \cite{Eigen2015,Wang2015}, image captioning \cite{Donahue2017}, etc.,  researchers have begun to utilize DCNN for low-level tasks such as edge detection \cite{Xie2015,Yang2016,YupeiWang2017,Liu2017,Maninis2017,Yu2017}. For example,  Xie et al. \cite{Xie2015} developed a HED network built upon the VGG-16 network \cite{Simonyan2014} which hierarchically obtains edge images at multiple scales. Edges obtained from the initial levels are more localized while those from the deeper levels are more global. The final edge is a linear combination of all edge images at different scales. Later, Kokkinos et al. \cite{Kokkinos2016} explicitly applied HED \cite{Xie2015} on the image pyramid. Yang et al. \cite{Yang2016} developed a fully convolutional encoder-decoder network (CEDN) similar to Noh et al. \cite{Noh2015}. 
The main drawback of these approaches is that the salient edges are obtained at the deeper layers with relatively lower resolution. Thus, the upsampled edge image tends to be blurry and less localized.

Several researchers also proposed the use of a refinement network for edge images at different hierarchies to achieve better edge detection results. For instance, Wang et al. \cite{YupeiWang2017} proposed a refinement module that fuses a top-down feature map from the backward pathway with the feature map from the current layer in the
forward pathway, and further up-samples the map by a small factor of two, which is then passed down the pathway. Liu et al. \cite{Liu2017} designed another type of refinement module, which uses all convolution layers at the same hierarchy to predict edge image at that level, to achieve a similar goal.

In this paper, we propose a novel Recursive Encoder-Decoder Network with Skip Connections (RED-NET) for edge detection in natural images. The proposed network is a novel integration of a Recursive Neural Network with an Encoder-Decoder architecture. Our encoder-decoder network is formed by DenseNet blocks \cite{Huang2017}, which are used to alleviate the vanishing-gradient problem, strengthen feature propagation and encourage feature reuse. The encoder network performs convolutions and poolings to produce a set of feature maps of different visual levels. The deeper the layer is, the higher level, more abstract, and less localized the features are. The encoder tends to learn more global and high-level features, and could ignore some finer information. The decoder network, which is topologically symmetric with the encoder network, first upsamples the feature maps by transposed convolutions (i.e. deconvolutions) followed by convolutions and finally returns the edge image with the same size as the input image. Nevertheless, since information related to finer details might be lost during the encoding stage, the decoded outputs are generally less detailed. As a result, edges generated by the encoder-dencoder network are usually blurry and less localized \cite{Yang2016}.

To overcome this limitation, in this paper, we propose to add \emph{skip-connections} \cite{Shi2016} that connect one layer in the encoder to the corresponding layer in the decoder of the same level of hierarchy. Since features from the early encoder are forwarded to the later decoder, skip-connections provide sharper visual details. Skip-connections have been widely used in deep learning community such as U-Net \cite{Ronneberger2015}, Deep Reflectane Map (DRM) \cite{Rematas2016}, ResNet \cite{He2016} and DenseNet \cite{Huang2017}. According to \cite{Shi2016,Huang2017}, skip-connections greatly improve gradient flow by allowing more even weight update in all of the layers.

We further enhance the network by adding a \emph{feedback loop} between the output edge map and the input \cite{Peng2016,Kim2016}. The purpose of this is to enable iterative refinement of the edges using a single network model. Increasing recursion depth can improve performance without introducing new parameters for additional convolutions and deconvolutions. The whole network can be modeled jointly with shared parameters and optimized in an end-to-end manner.

Furthermore, in order to force the network to learn more salient edges, we propose a simple but very effective data augmentation scheme by conducting random Gaussian blur to the input images. This also helps to reduce potential over-fitting as the input images is augmented randomly in each iteration during training.

In summary, the main contribution of this paper is to improve the deep learning algorithms used for edge detection by combining skip-connections and feedback loop into an encoder-decoder network, as well as a simple and effective Gaussian blurring based data augmentation. To the best of our knowledge, we are the first group applying the recursive network in low-level tasks such as edge detection. Our RED-NET experimentally demonstrates state-of-the-art results on popular boundary detection datasets including BSDS500 \cite{Arbelaez2011}, NYUD \cite{Silberman2012} and Pascal Context \cite{Mottaghi2014}.

\section{Related Work}
The literature of edge detection is very expansive. 
We will only be able to highlight a few representative works that 
are closely related to our work.

The early pioneering edge detection methods (e.g. \cite{Sobel1968,Duda1973,Fram1975,Marr1980,Torre1986,Canny1986,Perona1990,Freeman1991}) focused on low-level cues such as image intensity or color gradients. A complete overview of various low-level edge detectors can be found in \cite{Bowyer1999,Ziou1998}. For example, the well-known Canny edge detector \cite{Canny1986} finds the peak gradient orthogonal to edge direction.  In general, these low-level edge detectors are not very robust and may generate many false positives
or false negatives 

In the past decade, people have explored machine learning techniques for more accurate edge detection especially under more challenging conditions \cite{Martin2004,Dollar2006,Zheng2007,Mairal2008,Kokkinos2010,Arbelaez2011,Widynski2012,Leordeanu2014}. For example, Dollar et al. \cite{Dollar2006} used a boosted classifier to independently label each pixel using its surrounding image patch as input. Zheng et al. \cite{Zheng2007} combined low, mid, and high-level cues to achieve improved results for object-specific edge detection. Arbelaez et al. \cite{Arbelaez2011} combined multiple local cues into a global framework based on spectral clustering. Ren and Bo \cite{Ren2012} further improved the method of \cite{Arbelaez2011} by computing gradients across learned sparse codes of patch gradients. Lim et al. \cite{Lim2013} proposed an edge detection approach that classifies edge patches into sketch tokens using random forest classifiers. Sketch tokens are learned using supervised mid-level information in the form of hand drawn contours in images. Dollar et al. \cite{Dollar2015} learned more subtle variations in edge structure and lead to a more accurate and efficient algorithm. This structured edge detection method was considered one of the best method for edge detection thanks to its state-of-the-art performance and relatively fast speed.

Recently deep learning approaches become very popular and researchers have attempted to deploy it to edge detection. It is widely believed that accurate detection of edges requires object-level understanding of the image, an area in which deep learning is best known for. Kivinen et al. \cite{Kivinen2014} applied mean-and-covariance restricted Boltzmann machine (mcRBM) architecture \cite{Dahl2010} to edge detection and obtained competitive results. Starting from candidate contour points from Canny edge detector \cite{Canny1986}, DeepEdge \cite{Bertasius2015} extracts patches at four different scales and simultaneously run them through the five convolutional layers of the AlexNet \cite{Krizhevsky2012}. These convolutional layers are connected to two separately-trained network branches. The first branch is trained for classification, while the second branch is trained as a regressor. At testing time, the scalar outputs from these two sub-networks are averaged to produce the final score. DeepContour \cite{Shen2015} classified image patch of size $45 \times 45$ into background or one of the clustered shape classes by a 6-layer convolutional neural network. The disadvantage of both DeepEdge and DeepContour is that at testing time, it operates on the input image in a sliding window fashion (due to the fully-connected layers), which restricts the receptive size of the network to only a small image patch and thus may lose global information.

Inspired from FCN \cite{Long2015}, Xie et al. \cite{Xie2015,Xie2017} proposed the HED network which can be trained in an end-to-end manner. An interesting idea of this work is that the final edge map is fused from multiple edge maps obtained at different scales. The multi-scale edge maps are side outputs of a VGG-16 network \cite{Simonyan2014} and hence the shallower edge maps give finer detail edges while the deeper ones capture the more salient edges. The final result is linearly combined from all edge maps at multiple scales. The main drawback of this network is that salient edges are typically learned in the deeper layers, hence they are of low-quality when being up-sampled - edges are blurry and do not stick to actual image boundaries. Later, Kokkinos \cite{Kokkinos2016} proposed the Deep-Boundaries network, which is essentially a multi-scale HED \cite{Xie2015}. As being claimed by Kokkinos \cite{Kokkinos2016}, the explicit use of multiple scale improves the accuracy of edge detection. However, because being built upon the HED \cite{Xie2015} and fed by down-sampled images, Deep-Boundaries also suffers from the same issue as the HED.

To solve the issue of low quality salient edges, Wang et al. \cite{YupeiWang2017} and Liu et al. \cite{Liu2017} proposed the CED and RCF, respectively. Both papers proposed an extra network to synthesize the high resolution edge maps from low resolution ones instead of trivially using bilinear interpolation. For example, CED's refinement module fuses a top-down feature map from the backward pathway with the feature map from current layer in the forward pathway, and further up-samples the map by a small factor ($2\times$), which is then passed down the pathway.

Maninis et al. \cite{Maninis2017} proposed the Convolutional Oriented Boundaries (COB) which demonstrated state-of-the-art performance in edge detection. From a single pass of a base convolutional neural network, COB obtains multiscale oriented contours, combines them to build Ultrametric Contour Maps at different scales and finally fuses them into a single hierarchical segmentation structure.

Our RED-NET architecture is based on an encoder-decoder network with significant improvements. Firstly, we use DenseNet blocks within each convolution group. Secondly, we add skip-connections between encoder and decoder, which helps the gradient to more easily reach all the deep layers of a network. Additionally, finer details in the early stage of the encoder are preserved to be used in the decoder. Thirdly, our recursive network is used with convolutions to further increase the network depth with the same number of parameters. In the next section, we will describe our network architecture in depth followed by evaluation results.

\begin{figure*}[t]
\centering
\includegraphics[width=\textwidth]{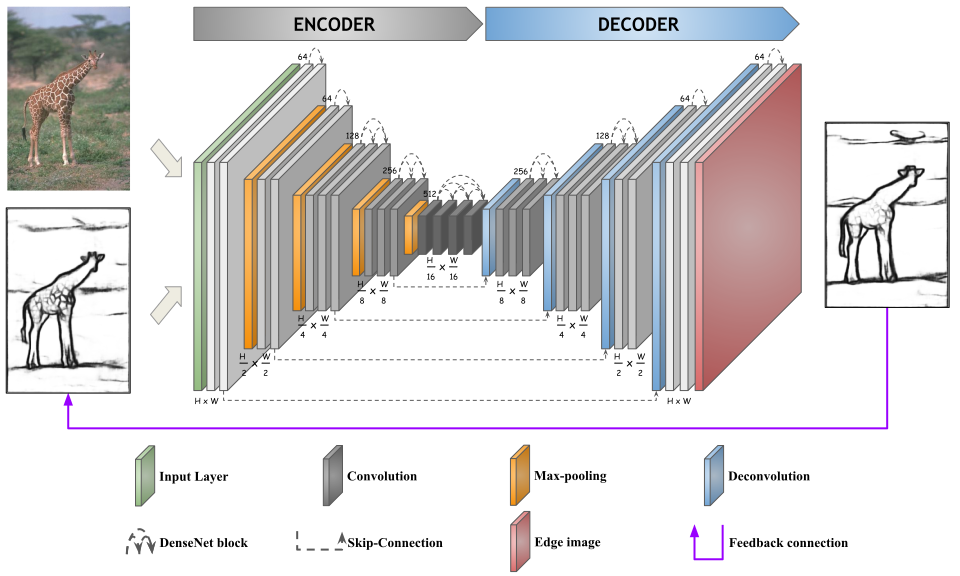}
\caption{The architecture of our Recursive Encoder-Decoder Network with Skip Connections (RED-NET). Encoder-decoder network is at the heart of our design which consists of DenseNet blocks. There are four skip-connections that connects one layer of the encoder to a corresponding layer of the decoder. The feedback connection enables a deeper network with no extra parameters.}
\label{fig:RED-NET}
\end{figure*}

\section{Recursive Encoder--Decoder Network with Skip-Connections}
Fig. \ref{fig:RED-NET} shows the architecture of our RED-NET. Our network takes as input an RGB image and a recursive edge image
, concatenates them (in the depth channel) and passes through an encoder-decoder network. The encoder consists of $5$ blocks of DenseNet \cite{Huang2017}. The decoder is symmetric with the encoder with max-pooling replaced by transposed convolution (i.e. deconvolution). Skip-connections connects corresponding layers of encoder and decoder at the same hierarchy. The decoder outputs an edge image of the same resolution as the input image, which serves as a recursive input to replace the edge image in the network (feedback loop). There are $L$ iterations and $L = 0$ indicates no feedback loop at all. In contrast to DeepEdge \cite{Bertasius2015} and DeepContour \cite{Shen2015} which can only be applied on image patch of fixed size due to the use of fully-connected layers, our RED-NET does not contain any fully-connected layer, and can consume images of any size. In the following sections, we elaborate the RED-NET in more details and discuss the training and testing procedures.

\subsection{Training Formulation}
\label{sec:training-formulation}
We denote our input training dataset by $S = \left\lbrace (X_i, Y_i) \right\rbrace_{i = 1}^N$ where $X_i$ denotes raw input image patch (we use patch size of $256 \times 256$ during all experiments) and $Y_i$ denotes the corresponding binary ground truth edge map for image patch $X_i$. The goal of the network is to produce edge maps approaching the ground truth. Let $\mathbf{W}$ be the collection of all network parameters for simplicity. The network runs through $L$ iterations, each of which produces an edge map $f^{(l)}(X_i | \mathbf{W})$ ($l=0,\dotsc,L$). Thus, $f^{(L)}(X_i | \mathbf{W})$ is the final output of the RED-NET. Consequently, the ultimate goal is to minimize the loss between the final edge map and the ground truth, or
\begin{equation}
\min_{\mathbf{W}} \quad \mathcal{L} \left( f^{(L)}(X_i | \mathbf{W}), Y_i \right)
\end{equation}
where $\mathcal{L}$ is the loss function, a weighted cross-entropy that will be discussed later.

Nevertheless, training such a deep network is not trivial when $L \ge 1$. Adapting the idea of deeply supervised network training \cite{Xie2015,Xie2017}, we also regularize the network by adding multiple losses for all $f^{(l)}(X_i | \mathbf{W})$. The goal now is to minimize the following.
\begin{equation}
\min_{\mathbf{W}} \quad \sum_{l = 0}^L \alpha_l \mathcal{L} \left( f^{(l)}(X_i | \mathbf{W}), Y_i \right)
\end{equation}
where $\left\lbrace \alpha_l \right\rbrace_{l = 0}^L$ are weights for edge maps at each iteration. We set $\alpha_l = l + 1$ to force the network to focus on the edge maps at later iterations.

\subsection{Testing Formulation}
\label{sec:testing-formulation}
During testing, given image $X$, we obtain the edge map predictions at all iterations of RED-NET, i.e. $f^{(l)}(X | \mathbf{W}), l = 0, \dots, L$. The final edge map is defined as the last one.
\begin{equation}
\hat{Y}_\text{RED} = f^{(L)}(X_i | \mathbf{W})
\label{eqn:final-edge-map-1}
\end{equation}
Alternatively, one may define the final edge map as a weighted combination of all edge maps with learnable weights $\gamma$ as follows.
\begin{equation}
\hat{Y}_\text{RED-NET} = \frac{\sum_{l = 0}^L \gamma_l \mathcal{L} \left( f^{(l)}(X_i | \mathbf{W}), Y_i \right)}{\sum_{l = 0}^L \gamma_l}
\label{eqn:final-edge-map-2}
\end{equation}
Empirically, when the network is trained properly, we do not notice significant difference between these two formulations (\ref{eqn:final-edge-map-1}) and (\ref{eqn:final-edge-map-2}) both visually and quantitatively. Therefore, we opt to use (\ref{eqn:final-edge-map-1}) for simplicity.

\begin{figure*}[!t]
\centering
\begin{subfigure}[b]{0.03\textwidth}
\rotatebox{90}{\hspace{0.5cm} \textbf{Original}}
\end{subfigure}
\begin{subfigure}[b]{0.22\textwidth}
\centering
\includegraphics[width = \textwidth]{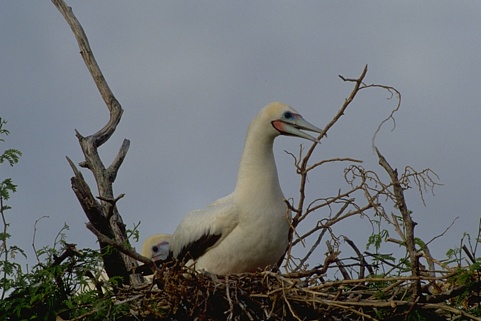}
\end{subfigure}
\begin{subfigure}[b]{0.22\textwidth}
\centering
\includegraphics[width = \textwidth]{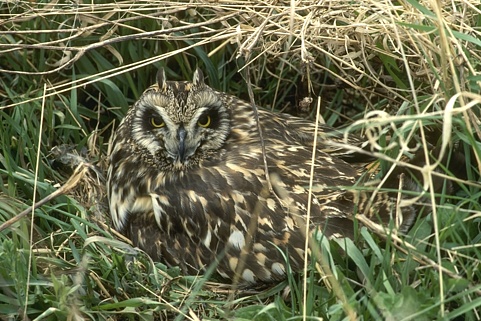}
\end{subfigure}
\begin{subfigure}[b]{0.22\textwidth}
\centering
\includegraphics[width = \textwidth]{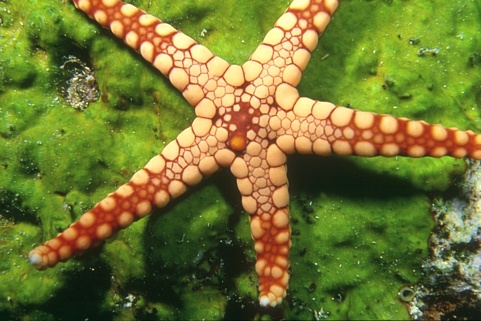}
\end{subfigure}
\begin{subfigure}[b]{0.22\textwidth}
\centering
\includegraphics[width = \textwidth]{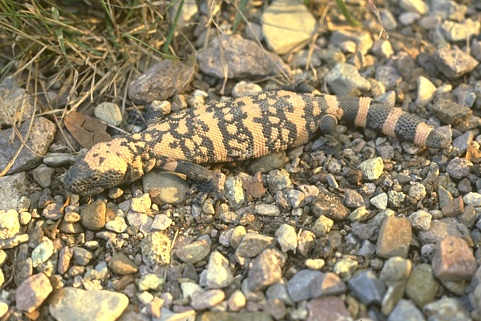}
\end{subfigure}

\vspace{2pt}

\begin{subfigure}[b]{0.03\textwidth}
\rotatebox{90}{\hspace{0.2cm} \textbf{Augmented}}
\end{subfigure}
\begin{subfigure}[b]{0.22\textwidth}
\centering
\includegraphics[width = \textwidth]{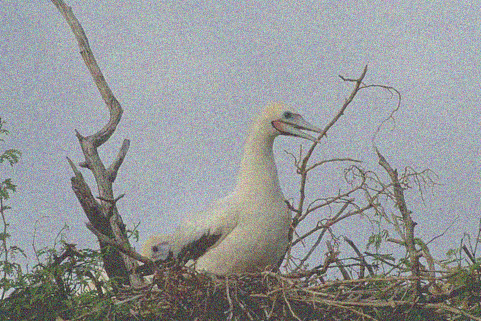}
\end{subfigure}
\begin{subfigure}[b]{0.22\textwidth}
\centering
\includegraphics[width = \textwidth]{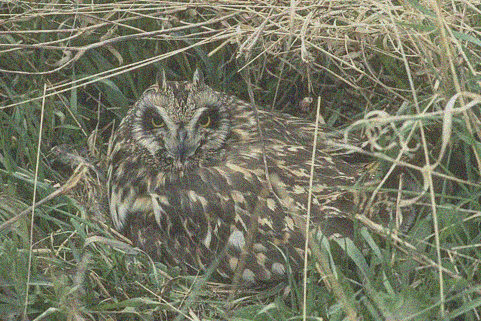}
\end{subfigure}
\begin{subfigure}[b]{0.22\textwidth}
\centering
\includegraphics[width = \textwidth]{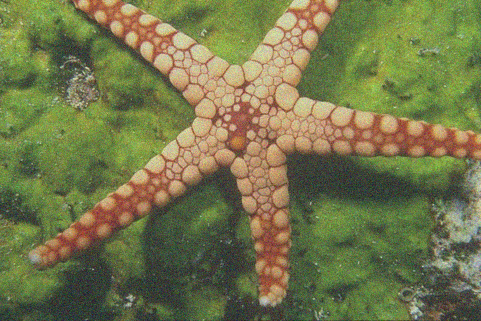}
\end{subfigure}
\begin{subfigure}[b]{0.22\textwidth}
\centering
\includegraphics[width = \textwidth]{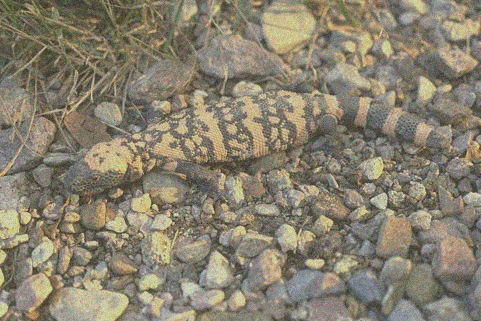}
\end{subfigure}

\caption{Original images (top row) are augmented with Gaussian noise to force the network to extract stronger edges.}
\label{fig:augmentation}
\end{figure*}

\subsection{Network Architecture}
\label{sec:network-architecture}
\subsubsection{Encoder}
The encoder extracts features from input image, so we need an architecture that is deep and can efficiently generate perceptually multi-level features. Inspired from the recent success of DenseNet \cite{Huang2017} on image classification, we design our encoder by stacking $5$ DenseNet blocks. The first block consists of two $5 \times 5$ convolution layers with $64$ kernels for each, followed by a similar second block with max-pooling layer in between which downsamples the feature maps and hence forces the network to learn good global features. Starting from the third block, we double the number of kernels for each successive block, which results in a $512$-dimension feature maps after the fifth block. Moreover, we also increase the number of convolution layers to $3$, $3$ and $4$ for the third, fourth and fifth blocks, respectively for more powerful architecture. Every convolution layer in the encoder composes of a convolution layer, a batch normalization layer \cite{Ioffe2015} and a leaky rectified unit activation \cite{Maas2013} (with leaking coefficient of $0.1$) in this order.

\subsubsection{Decoder}
The decoder maps the learned features to another space and eventually reaches the edge image. This network is symmetric with the encoder with $5$ DenseNet blocks. We use transposed convolutions (or deconvolutions) to upsample the feature maps corresponding to max-pooling in the encoder. The transposed convolutions are initialized as bilinear filters which purely serve as upsample filters. At the last layer, the decoder returns an edge prediction from the $64$-channel layer via convolution. To facilitate the training, we also use batch normalization and leaky rectified unit in the same way as in the encoder except for the last layer which only consists of a convolution followed by a sigmoid activation.

\subsubsection{Skip-connections}
The encoder progressively extracts and down-samples features, while the decoder upsamples and combines them to construct the output. The sizes of feature maps are exactly mirrored in our network. We concatenate early encoded features (from the encoder) to the corresponding decoded features (from the decoder) at the same spatial resolution, in order to obtain local sharp details preserved in early encoder layers. There are four of such skip-connections corresponding to four different level of hierarchies, which are called mirror-links. Mirror-link is a form of skip connection which has been proven effective in many deep network such as ResNet \cite{He2016} and DenseNet \cite{Huang2017}. Besides the sharpness, these skip-connections could also regulate gradient flow and allow better trained network.

\subsubsection{Feedback loop}
This is a recursive connection similar to the Recurrent Neural Network (RNN). In contrast to RNN in which recurrence targets temporal sequence and tries to learn temporal changes, our feedback loop refines the edge map progressively without introducing more network parameters. At the beginning, there is no edge image generated, so the initial edge image is set as a blank image (i.e. zero-image). After the first pass through the encoder-decoder network, the output edge map is recursively fed back to the input and repeatedly processed through the shared encoder-decoder network. The whole RED-NET is jointly optimized in an end-to-end manner. Due to memory limit, we only conduct experiments for $L = 2$.

\subsection{Loss function}
We use weighted sigmoid cross-entropy function to compute the loss between our predicted edge $\hat{Y}_\text{RED-NET}$ (or other intermediate edge images $f^{(l)}(X_i | \mathbf{W})$, $l=0,\dotsc,L$) and the ground truth edge image $Y$ as follows.
\begin{equation}
\mathcal{L} \left( \hat{Y}_\text{RED-NET}, Y \right) = -(1 - \beta) \sum_{j \in Y_+} \log \hat{Y}_\text{RED-NET} - \beta \sum_{j in Y_-} \log \left( 1 - \hat{Y}_\text{RED-NET} \right)
\label{eqn:loss}
\end{equation}
where $Y_+$ and $Y_-$ denote edge and non-edge pixels, respectively and $\beta = \frac{|Y_+|}{|Y|}$ to balance the relative importance of these two classes.

\subsection{Implementation}
\label{sec:implementation}
We implement our framework using the publicly available TensorFlow \cite{tensorflow2015}.

\subsubsection{Hyper-parameters}
In contrast to fine-tuning CNN for image classification, adapting CNN for pixel-wise output requires special care. Even with the proper initialization or a pre-trained model, sparse ground truth distributions coupled with conventional loss functions lead to difficulties in network convergence. Through experimentation, we choose the following hyper-parameters: mini-batch size of $8$, convolutional filters randomly initialized by Gaussian distribution with zero-mean and standard deviation of $0.01$, convolutional biases all zero-initialized, deconvolutions initialized as bilinear filters, weight decay of $10^{-6}$, training epochs equal $\num{500}$. Furthermore, we use Adam optimizer \cite{Kingma2015} with initial learning rate $10^{-4}$. As mention earlier, we extract image patches of size $256 \times 256$ for training but use the whole image during testing.

\subsubsection{Data augmentation}
\begin{figure*}[!t]
\centering
\begin{subfigure}[b]{0.04\textwidth}
\rotatebox{90}{\hspace{0.43cm} \textbf{Input Image}}
\end{subfigure}
\begin{subfigure}[b]{0.223\textwidth}
\centering
\includegraphics[width = \textwidth]{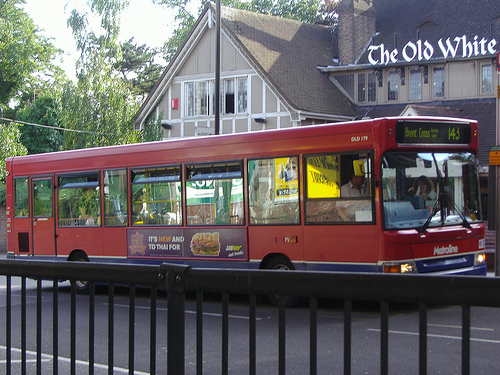}
\end{subfigure}
\begin{subfigure}[b]{0.223\textwidth}
\centering
\includegraphics[width = \textwidth]{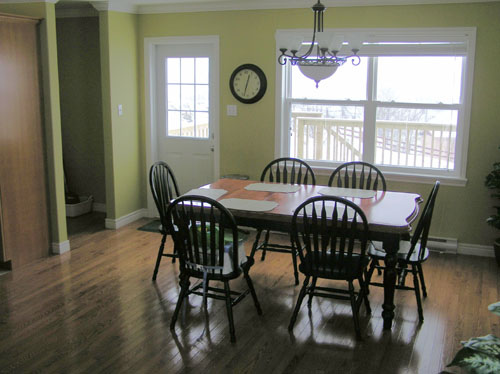}
\end{subfigure}
\begin{subfigure}[b]{0.223\textwidth}
\centering
\includegraphics[width = \textwidth]{figures/get-gt/2008_000112-image.png}
\end{subfigure}
\begin{subfigure}[b]{0.223\textwidth}
\centering
\includegraphics[width = \textwidth]{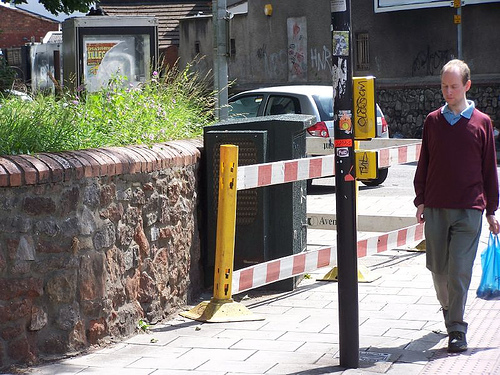}
\end{subfigure}

\vspace{2pt}

\begin{subfigure}[b]{0.04\textwidth}
\rotatebox{90}{\hspace{0.6cm} \textbf{Semantic}}
\rotatebox{90}{\hspace{0.3cm} \textbf{Segmentation}}
\end{subfigure}
\begin{subfigure}[b]{0.223\textwidth}
\centering
\includegraphics[width = \textwidth]{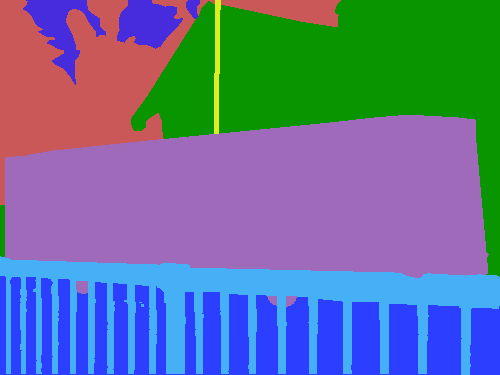}
\end{subfigure}
\begin{subfigure}[b]{0.223\textwidth}
\centering
\includegraphics[width = \textwidth]{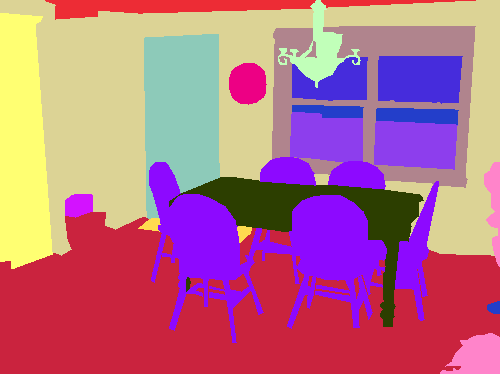}
\end{subfigure}
\begin{subfigure}[b]{0.223\textwidth}
\centering
\includegraphics[width = \textwidth]{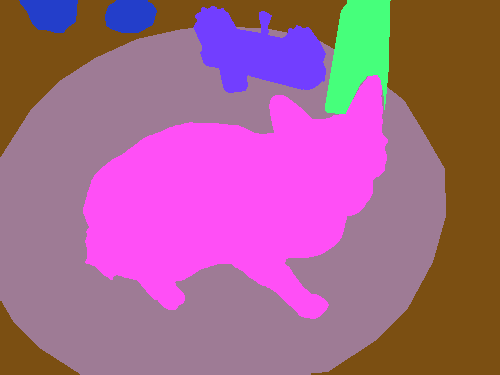}
\end{subfigure}
\begin{subfigure}[b]{0.223\textwidth}
\centering
\includegraphics[width = \textwidth]{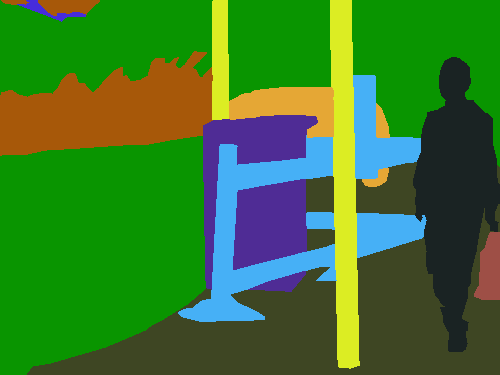}
\end{subfigure}

\vspace{2pt}

\begin{subfigure}[b]{0.04\textwidth}
\rotatebox{90}{\hspace{0.15cm} \textbf{Boundary Pixels}}
\end{subfigure}
\begin{subfigure}[b]{0.223\textwidth}
\centering
\frame{\includegraphics[width = \textwidth]{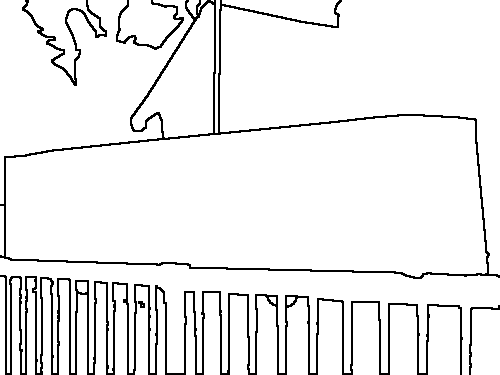}}
\end{subfigure}
\begin{subfigure}[b]{0.223\textwidth}
\centering
\frame{\includegraphics[width = \textwidth]{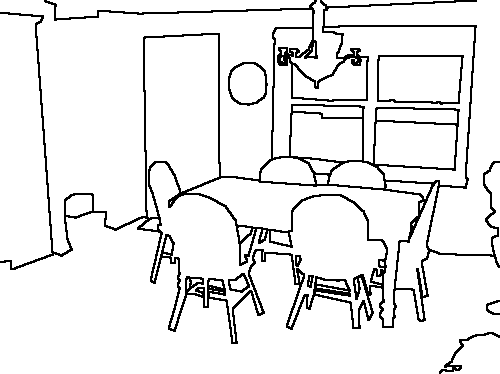}}
\end{subfigure}
\begin{subfigure}[b]{0.223\textwidth}
\centering
\frame{\includegraphics[width = \textwidth]{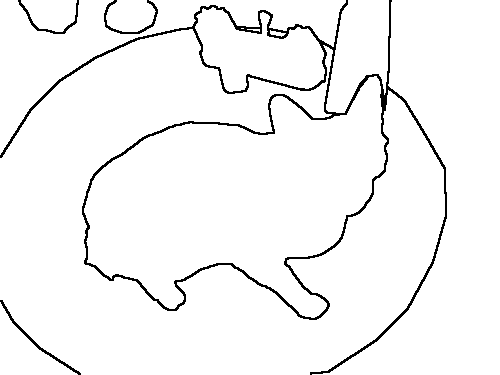}}
\end{subfigure}
\begin{subfigure}[b]{0.223\textwidth}
\centering
\frame{\includegraphics[width = \textwidth]{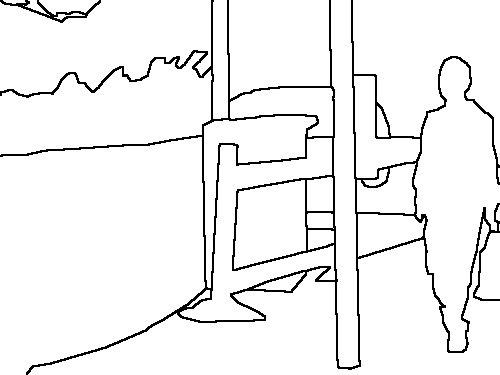}}
\end{subfigure}

\vspace{2pt}

\begin{subfigure}[b]{0.04\textwidth}
\rotatebox{90}{\hspace{0.6cm}\textbf{Edge Image}}
\end{subfigure}
\begin{subfigure}[b]{0.223\textwidth}
\centering
\frame{\includegraphics[width = \textwidth]{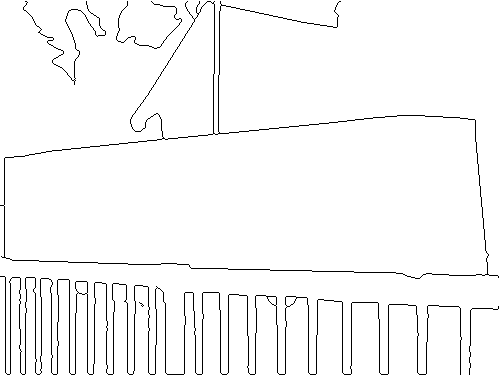}}
\end{subfigure}
\begin{subfigure}[b]{0.223\textwidth}
\centering
\frame{\includegraphics[width = \textwidth]{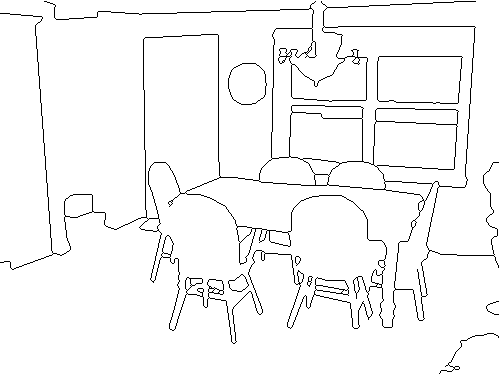}}
\end{subfigure}
\begin{subfigure}[b]{0.223\textwidth}
\centering
\frame{\includegraphics[width = \textwidth]{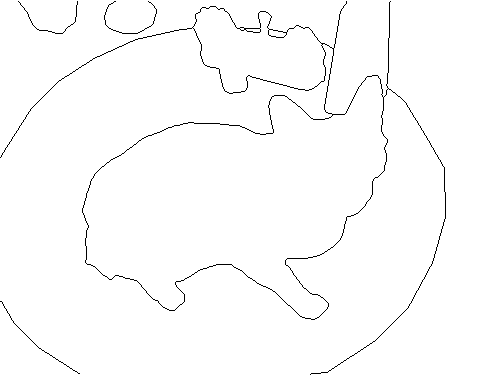}}
\end{subfigure}
\begin{subfigure}[b]{0.223\textwidth}
\centering
\frame{\includegraphics[width = \textwidth]{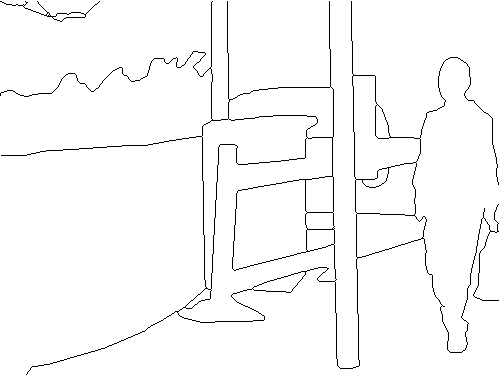}}
\end{subfigure}

\caption{Ground truth edge image generation. From an input image (first row) with ground truth semantic segmentation (second row), we identify all boundary pixels (third row) and then apply image thinning (e.g. \texttt{MATLAB}'s \texttt{bwmorph}) to obtain the ground truth edge image (last row).}
\label{fig:how-to-get-ground-truth}
\end{figure*}

Data augmentation has proven to be a crucial technique in training deep neural networks. For each training image, we randomly sample $500$ patches, each of size $256 \times 256$, which is a kind of image cropping. We further randomly flip the training image horizontally. These together lead to an augmented training set that is a factor of $500$ times larger than the unaugmented set.

In addition, we add random Gaussian noise (black-and-white noise) to the training images by sampling from a Gaussian distribution with zero-mean and standard deviation of $20$ (assuming image intensities are within $[0, 255]$) (Fig. \ref{fig:augmentation}). This data augmentation forces the network to learn the stronger edges such as object contours over finer texture ones. This augmentation also helps combat over-fitting because each training image is augmented differently in each iteration.

\subsubsection{Running time}
Training ranges from $4$ hours for the BSDS500 dataset with $300$ images to $50$ hours for the Pascal Context dataset with $\num{7605}$ images on a single Titan-X GPU. RED-NET produces an edge response for an image of size $512 \times 512$ in about $270$ milliseconds including interface overhead (e.g. image loading), which is approximately $3.4$ frames/second. This is significantly more efficient then existing CNNs such as DeepEdge \cite{Bertasius2015}, DeepContour \cite{Shen2015} and COB \cite{Maninis2017}.

\section{Evaluation}
This section presents the performance of our RED-NET on the well-known datasets for edge detection such as BSDS500 \cite{Arbelaez2011}, NYUD \cite{Silberman2012} and Pascal Context \cite{Mottaghi2014} (see Table \ref{tbl:datasets}). We adopt three standard evaluation metrics commonly used for edge detection, fixed contour threshold ODS F-score, per-image best threshold OIS F-score, and average precision AP \cite{Arbelaez2011}. We compare our method against popular state-of-the-art methods including both the non-deep learning and deep learning approaches. For a fair quantitative comparison, we apply a standard non-maximal suppression technique \cite{Dollar2015} to all edge maps generated by all methods to obtain thinned edges before evaluation.

\subsection{Datasets}
\begin{table}[!t]
\centering
\caption{Datasets and Parameters}
\label{tbl:datasets}
\begin{tabular}{rR{1.3cm}R{1.3cm}R{1.8cm}}
\toprule
\textbf{Dataset} & \textbf{\# train} & \textbf{\# test} & \textbf{\texttt{maxDist}}\\
\hline
BSDS500 \cite{Arbelaez2011} & 300 & 200 & 0.0075\\
NYUD-v2 \cite{Silberman2012} & 795 & 654 & 0.011\\
Pascal Context \cite{Mottaghi2014} & 7605 & 2498 & 0.0075\\
\bottomrule
\end{tabular}
\end{table}

\begin{table}[!t]
\centering
\caption{BSDS500 \cite{Arbelaez2011} test evaluation}
\label{tbl:bsds500}
\begin{tabular}{rR{1.4cm}R{1.4cm}R{1.4cm}}
\toprule
\textbf{Method} & \textbf{ODS} & \textbf{OIS} & \textbf{AP}\\
\hline
Canny \cite{Canny1986} & 0.600 & 0.640 & 0.580\\
MShift \cite{Comaniciu2002} & 0.601 & 0.644 & 0.493\\
EGB \cite{Felzenszwalb2004} & 0.610 & 0.640 & 0.560\\
ISCRA \cite{Ren2013} & 0.724 & 0.752 & 0.783\\
gPb-owt-ucm \cite{Arbelaez2011} & 0.726 & 0.757 & 0.696\\
Sketch Tokens \cite{Lim2013} & 0.727 & 0.746 & 0.780\\
SCG \cite{Ren2012} & 0.739 & 0.758 & 0.773\\
SE \cite{Dollar2015} & 0.746 & 0.767 & 0.803\\
OEF \cite{Hallman2015} & 0.749 & 0.772 & 0.817\\
MCG \cite{Arbelaez2014} & 0.747 & 0.779 & 0.759\\
LEP \cite{Zhao2015} & 0.757 & 0.793 & 0.828\\
DeepNets \cite{Kivinen2014} & 0.738 & 0.759 & 0.758\\
N4-Fields \cite{Ganin2015} & 0.753 & 0.769 & 0.784\\
DeepEdge \cite{Bertasius2015} & 0.753 & 0.772 & 0.807\\
CSCNN \cite{Hwang2015} & 0.756 & 0.775 & 0.798\\
DeepContour \cite{Shen2015} & 0.756 & 0.773 & 0.797\\
HED-fusion \cite{Xie2017} & 0.782 & 0.804 & 0.833\\
HED-late-merging \cite{Xie2017} & 0.788 & 0.808 & 0.840\\
CEDN \cite{Yang2016} & 0.788 & 0.804 & 0.834\\
COB \cite{Maninis2017} & 0.793 & 0.820 & 0.859\\
CED \cite{YupeiWang2017} & 0.803 & 0.820 & 0.871\\
RCF \cite{Liu2017} & 0.806 & 0.823 & --\\
\hline
RED-NET (ours) & 0.808 & 0.828 & 0.827\\
\bottomrule
\end{tabular}
\end{table}

\begin{table*}[!t]
\centering
\caption{NYUD-v2 \cite{Silberman2012} test evaluation}
\label{tbl:nyud}
\begin{tabular}{rR{1cm}R{1cm}R{1cm}}
\toprule
\textbf{Method} & \textbf{ODS} & \textbf{OIS} & \textbf{AP}\\
\hline
gPb-owt-ucm \cite{Arbelaez2011} & 0.726 & 0.757 & 0.696\\
Silberman et al. \cite{Silberman2012} & 0.658 & 0.661 & n/a\\
SE \cite{Dollar2015} & 0.685 & 0.699 & 0.679\\
MCG-B \cite{Arbelaez2014} & 0.652 & 0.681 & 0.613\\
HED-RGB \cite{Xie2017} & 0.720 & 0.734 & 0.734\\
HED-HHA \cite{Xie2017} & 0.682 & 0.695 & 0.702\\
HED-RGB-HHA \cite{Xie2017} & 0.746 & 0.761 & 0.786\\
ResNet50-RGB-HHA \cite{Maninis2017} & 0.745 & 0.762 & 0.792\\
ResNet50-RGB \cite{Maninis2017} & 0.746 & 0.761 & 0.789\\
ResNet50-RGBD \cite{Maninis2017} & 0.683 & 0.699 & 0.681\\
RCF-RGB \cite{Liu2017} & 0.729 & 0.742 & --\\
RCF-RGB-HHA \cite{Maninis2017} & 0.757 & 0.771 & --\\
COB-PC \cite{Maninis2017} & 0.710 & 0.735 & 0.734\\
COB-RGB \cite{Maninis2017} & 0.778 & 0.799 & 0.814\\
COB-RGB-HHA \cite{Maninis2017} & 0.784 & 0.805 & 0.825\\
\hline
RED-NET (ours) & 0.793 & 0.813 & 0.832\\
\bottomrule
\end{tabular}
\end{table*}

\begin{table*}[!t]
\centering
\caption{Pascal Context \cite{Mottaghi2014} test evaluation}
\label{tbl:pascal-context}
\begin{tabular}{rrrr}
\toprule
\textbf{Method} & \textbf{ODS} & \textbf{OIS} & \textbf{AP}\\
\hline
SE \cite{Dollar2015} & 0.533 & 0.568 & 0.496\\
LEP-B \cite{Zhao2015} & 0.570 & 0.636 & 0.547\\
MCG-B \cite{Arbelaez2014} & 0.554 & 0.609 & 0.528\\
HED \cite{Xie2015} & 0.688 & 0.707 & 0.704\\
CEDN \cite{Yang2016} & 0.702 & 0.718 & 0.744\\
COB \cite{Maninis2017} & 0.750 & 0.781 & 0.773\\
\hline
RED-NET ($L=0$) (no data aug.) & 0.744 & 0.769 & 0.771\\
RED-NET ($L=2$) (no data aug.) & 0.759 & 0.784 & 0.784\\
RED-NET ($L=2$) (with data aug.) & 0.761 & 0.785 & 0.787\\
\bottomrule
\end{tabular}
\end{table*}

\begin{table*}[!t]
\centering
\caption{Cross-dataset performance of RED-NET}
\label{tbl:cross-dataset}
\begin{tabular}{rr|rr|rrr}
\toprule
\textbf{Train} & & \textbf{Test} & & \textbf{ODS} & \textbf{OIS} & \textbf{AP}\\
\hline
\multirow{2}{*}{Pascal Context} & & BSDS500 & & 0.755 & 0.781 & 0.828\\
& & NYUD & & 0.732 & 0.766 & 0.783\\
\hline
\multirow{2}{*}{BSDS500} & & Pascal Context & & 0.643 & 0.653 & 0.681\\
& & NYUD & & 0.627 & 0.653 & 0.703\\
\bottomrule
\end{tabular}
\end{table*}

\begin{figure*}
\centering
\begin{subfigure}[b]{0.16\textwidth}
\centering
\textbf{Input}
\end{subfigure}
\begin{subfigure}[b]{0.16\textwidth}
\centering
\textbf{Human}
\end{subfigure}
\begin{subfigure}[b]{0.16\textwidth}
\centering
\textbf{RED-NET (ours)}
\end{subfigure}
\begin{subfigure}[b]{0.16\textwidth}
\centering
\textbf{HED \cite{Xie2015}}
\end{subfigure}
\begin{subfigure}[b]{0.16\textwidth}
\centering
\textbf{SE \cite{Dollar2015}}
\end{subfigure}
\begin{subfigure}[b]{0.16\textwidth}
\centering
\textbf{gPb \cite{Arbelaez2011}}
\end{subfigure}

\vspace{2pt}

\begin{subfigure}[b]{0.16\textwidth}
\centering
\includegraphics[width = \textwidth]{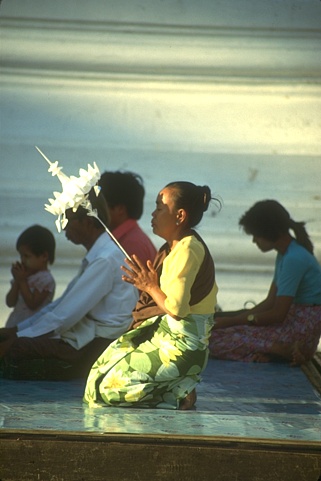}
\end{subfigure}
\begin{subfigure}[b]{0.16\textwidth}
\centering
\frame{\includegraphics[width = \textwidth]{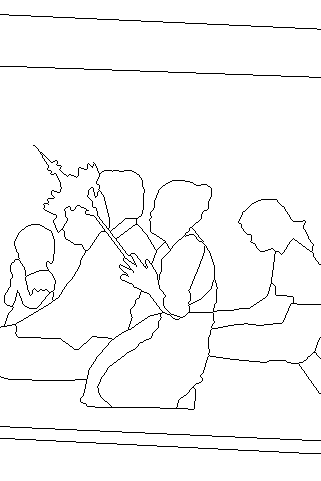}}
\end{subfigure}
\begin{subfigure}[b]{0.16\textwidth}
\centering
\frame{\includegraphics[width = \textwidth]{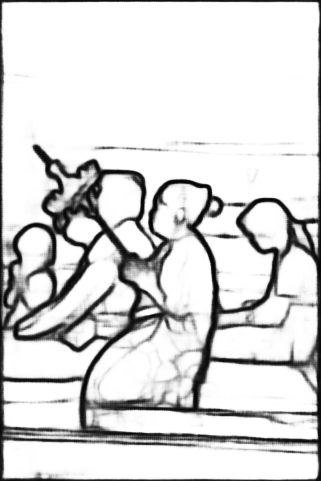}}
\end{subfigure}
\begin{subfigure}[b]{0.16\textwidth}
\centering
\frame{\includegraphics[width = \textwidth]{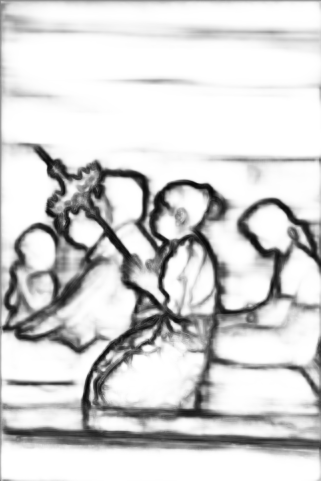}}
\end{subfigure}
\begin{subfigure}[b]{0.16\textwidth}
\centering
\frame{\includegraphics[width = \textwidth]{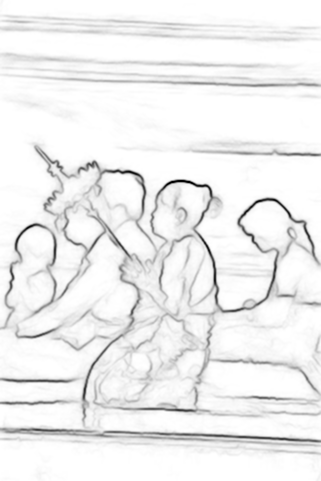}}
\end{subfigure}
\begin{subfigure}[b]{0.16\textwidth}
\centering
\frame{\includegraphics[width = \textwidth]{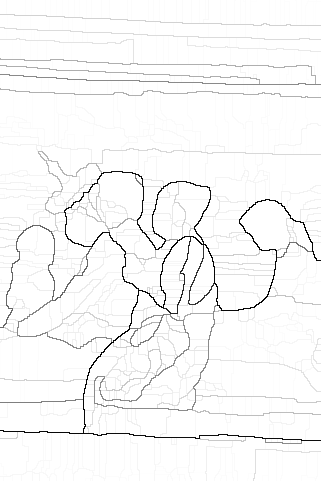}}
\end{subfigure}

\vspace{2pt}

\begin{subfigure}[b]{0.16\textwidth}
\centering
\includegraphics[width = \textwidth]{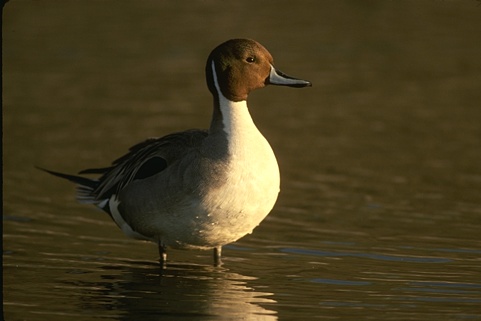}
\end{subfigure}
\begin{subfigure}[b]{0.16\textwidth}
\centering
\frame{\includegraphics[width = \textwidth]{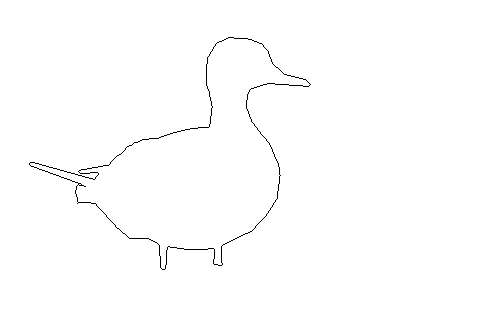}}
\end{subfigure}
\begin{subfigure}[b]{0.16\textwidth}
\centering
\frame{\includegraphics[width = \textwidth]{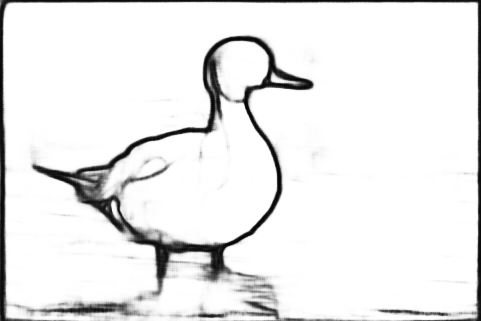}}
\end{subfigure}
\begin{subfigure}[b]{0.16\textwidth}
\centering
\frame{\includegraphics[width = \textwidth]{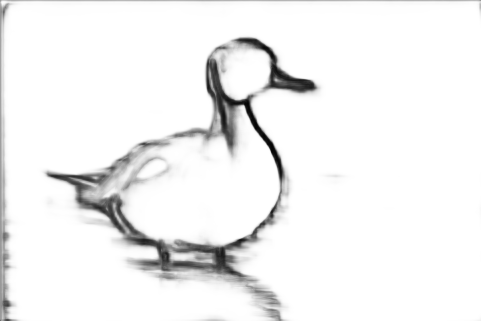}}
\end{subfigure}
\begin{subfigure}[b]{0.16\textwidth}
\centering
\frame{\includegraphics[width = \textwidth]{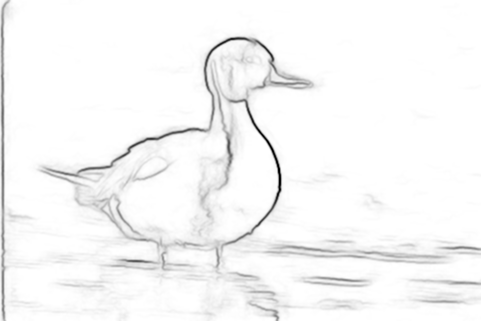}}
\end{subfigure}
\begin{subfigure}[b]{0.16\textwidth}
\centering
\frame{\includegraphics[width = \textwidth]{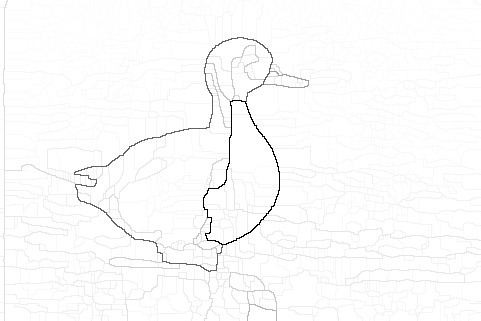}}
\end{subfigure}

\vspace{2pt}

\begin{subfigure}[b]{0.16\textwidth}
\centering
\includegraphics[width = \textwidth]{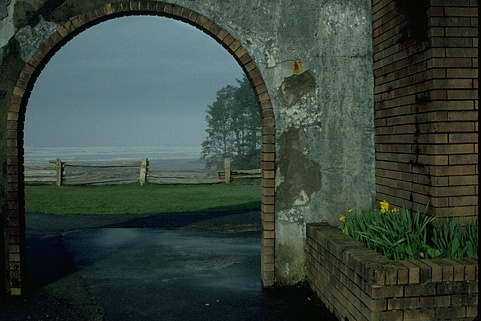}
\end{subfigure}
\begin{subfigure}[b]{0.16\textwidth}
\centering
\frame{\includegraphics[width = \textwidth]{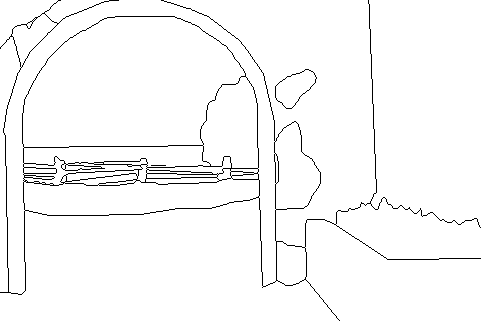}}
\end{subfigure}
\begin{subfigure}[b]{0.16\textwidth}
\centering
\frame{\includegraphics[width = \textwidth]{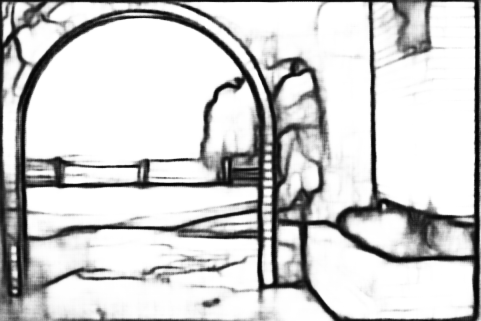}}
\end{subfigure}
\begin{subfigure}[b]{0.16\textwidth}
\centering
\frame{\includegraphics[width = \textwidth]{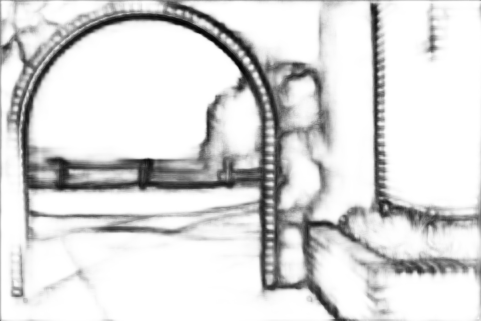}}
\end{subfigure}
\begin{subfigure}[b]{0.16\textwidth}
\centering
\frame{\includegraphics[width = \textwidth]{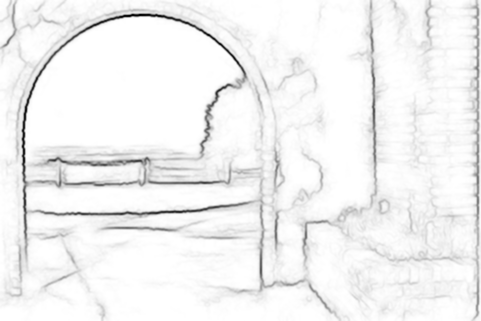}}
\end{subfigure}
\begin{subfigure}[b]{0.16\textwidth}
\centering
\frame{\includegraphics[width = \textwidth]{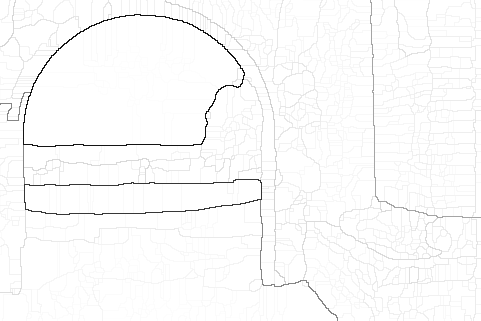}}
\end{subfigure}

\vspace{2pt}

\begin{subfigure}[b]{0.16\textwidth}
\centering
\includegraphics[width = \textwidth]{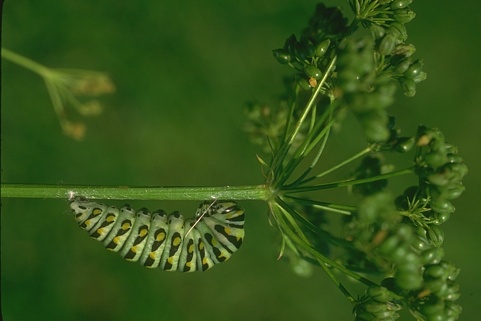}
\end{subfigure}
\begin{subfigure}[b]{0.16\textwidth}
\centering
\frame{\includegraphics[width = \textwidth]{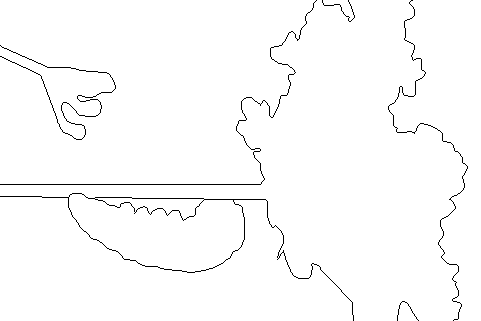}}
\end{subfigure}
\begin{subfigure}[b]{0.16\textwidth}
\centering
\frame{\includegraphics[width = \textwidth]{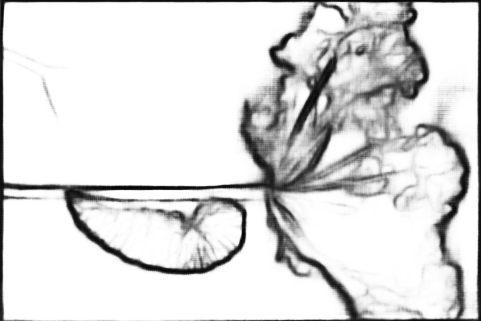}}
\end{subfigure}
\begin{subfigure}[b]{0.16\textwidth}
\centering
\frame{\includegraphics[width = \textwidth]{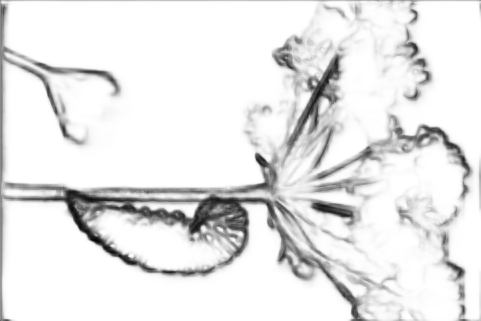}}
\end{subfigure}
\begin{subfigure}[b]{0.16\textwidth}
\centering
\frame{\includegraphics[width = \textwidth]{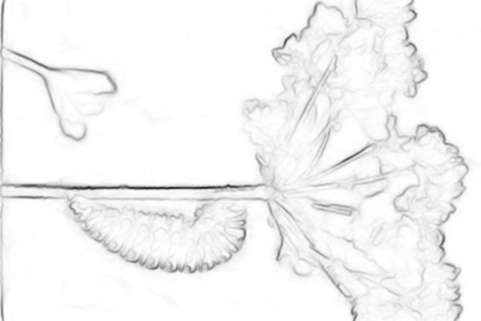}}
\end{subfigure}
\begin{subfigure}[b]{0.16\textwidth}
\centering
\frame{\includegraphics[width = \textwidth]{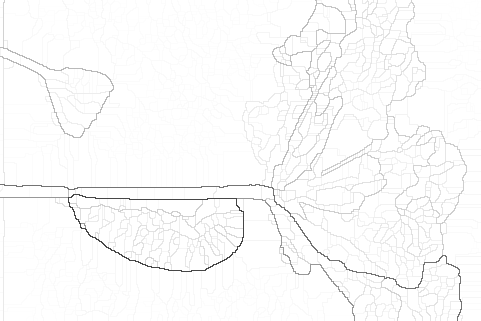}}
\end{subfigure}

\vspace{2pt}

\begin{subfigure}[b]{0.16\textwidth}
\centering
\includegraphics[width = \textwidth]{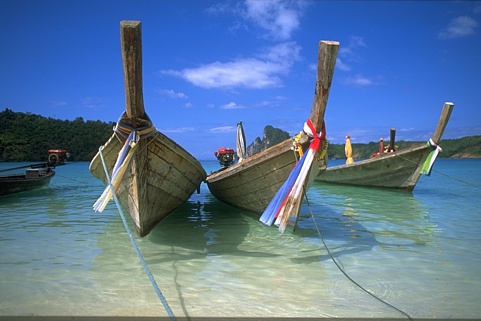}
\end{subfigure}
\begin{subfigure}[b]{0.16\textwidth}
\centering
\frame{\includegraphics[width = \textwidth]{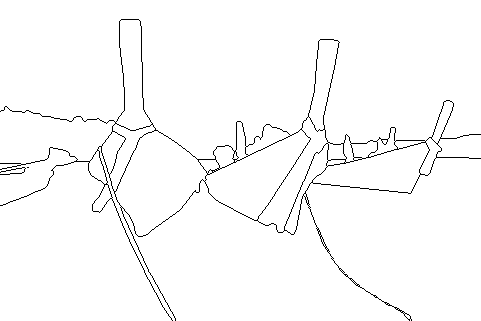}}
\end{subfigure}
\begin{subfigure}[b]{0.16\textwidth}
\centering
\frame{\includegraphics[width = \textwidth]{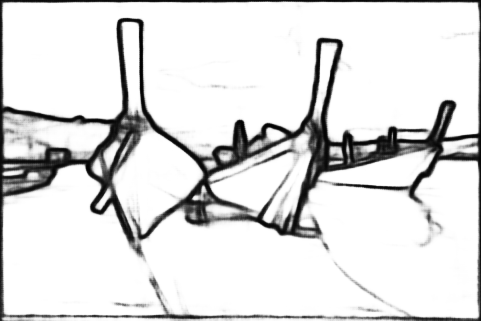}}
\end{subfigure}
\begin{subfigure}[b]{0.16\textwidth}
\centering
\frame{\includegraphics[width = \textwidth]{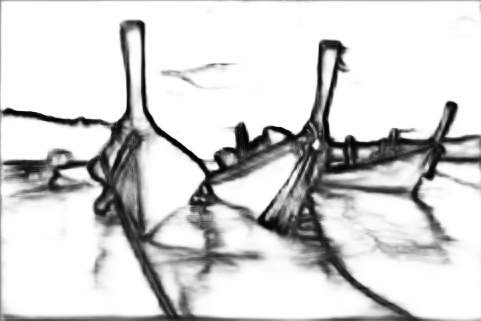}}
\end{subfigure}
\begin{subfigure}[b]{0.16\textwidth}
\centering
\frame{\includegraphics[width = \textwidth]{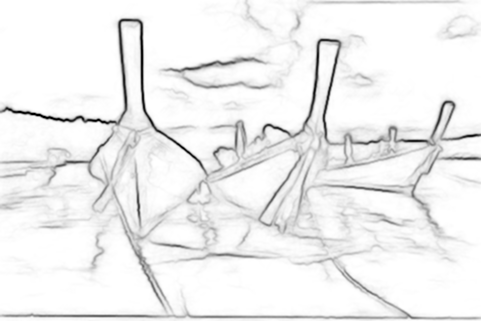}}
\end{subfigure}
\begin{subfigure}[b]{0.16\textwidth}
\centering
\frame{\includegraphics[width = \textwidth]{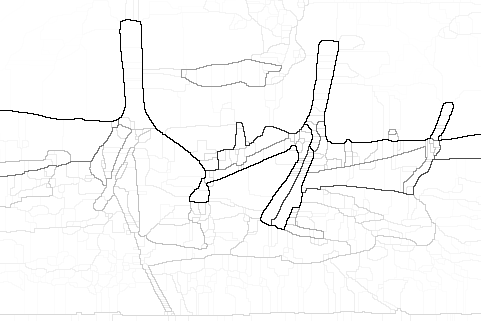}}
\end{subfigure}

\vspace{2pt}

\begin{subfigure}[b]{0.16\textwidth}
\centering
\includegraphics[width = \textwidth]{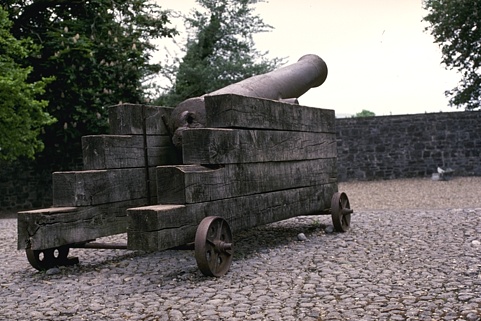}
\end{subfigure}
\begin{subfigure}[b]{0.16\textwidth}
\centering
\frame{\includegraphics[width = \textwidth]{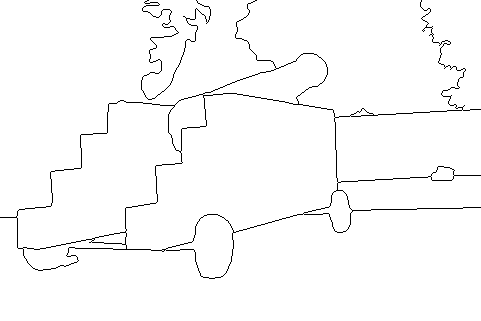}}
\end{subfigure}
\begin{subfigure}[b]{0.16\textwidth}
\centering
\frame{\includegraphics[width = \textwidth]{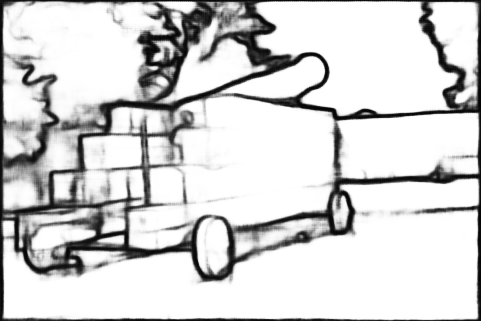}}
\end{subfigure}
\begin{subfigure}[b]{0.16\textwidth}
\centering
\frame{\includegraphics[width = \textwidth]{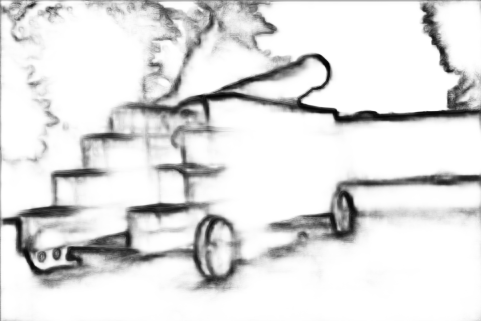}}
\end{subfigure}
\begin{subfigure}[b]{0.16\textwidth}
\centering
\frame{\includegraphics[width = \textwidth]{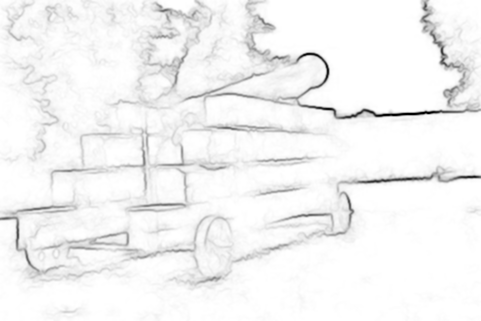}}
\end{subfigure}
\begin{subfigure}[b]{0.16\textwidth}
\centering
\frame{\includegraphics[width = \textwidth]{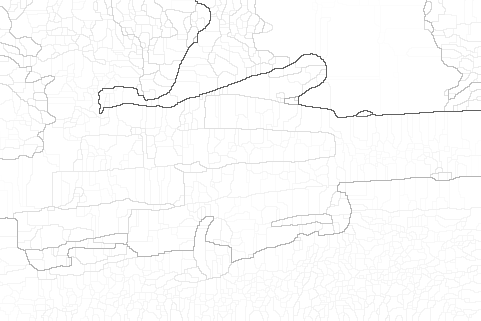}}
\end{subfigure}

\vspace{2pt}

\begin{subfigure}[b]{0.16\textwidth}
\centering
\includegraphics[width = \textwidth]{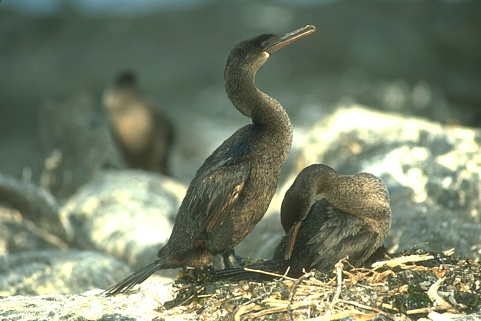}
\end{subfigure}
\begin{subfigure}[b]{0.16\textwidth}
\centering
\frame{\includegraphics[width = \textwidth]{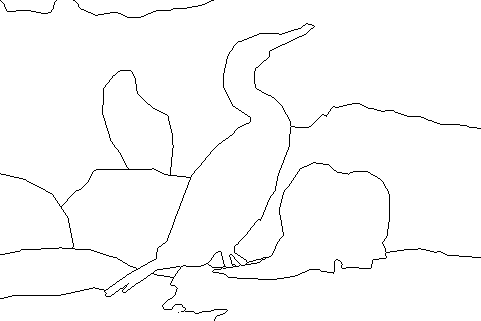}}
\end{subfigure}
\begin{subfigure}[b]{0.16\textwidth}
\centering
\frame{\includegraphics[width = \textwidth]{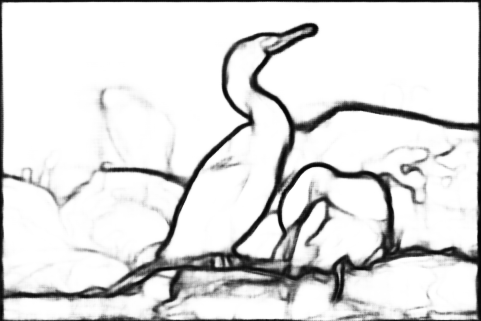}}
\end{subfigure}
\begin{subfigure}[b]{0.16\textwidth}
\centering
\frame{\includegraphics[width = \textwidth]{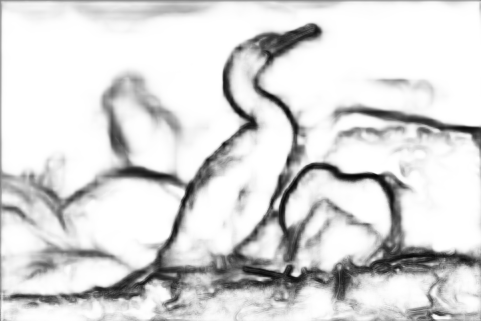}}
\end{subfigure}
\begin{subfigure}[b]{0.16\textwidth}
\centering
\frame{\includegraphics[width = \textwidth]{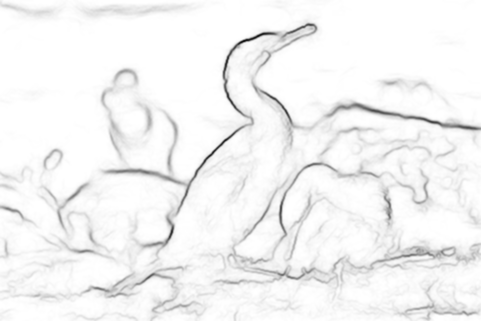}}
\end{subfigure}
\begin{subfigure}[b]{0.16\textwidth}
\centering
\frame{\includegraphics[width = \textwidth]{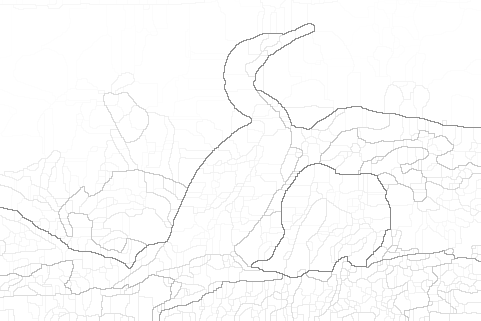}}
\end{subfigure}

\vspace{2pt}

\begin{subfigure}[b]{0.16\textwidth}
\centering
\includegraphics[width = \textwidth]{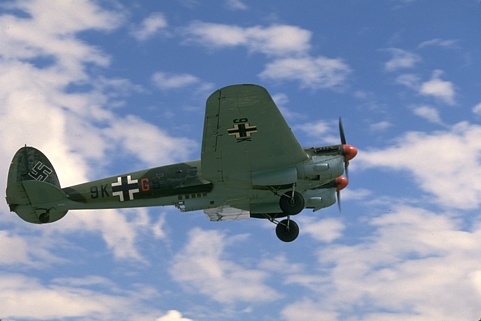}
\end{subfigure}
\begin{subfigure}[b]{0.16\textwidth}
\centering
\frame{\includegraphics[width = \textwidth]{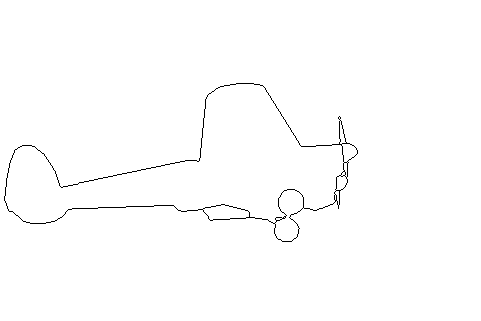}}
\end{subfigure}
\begin{subfigure}[b]{0.16\textwidth}
\centering
\frame{\includegraphics[width = \textwidth]{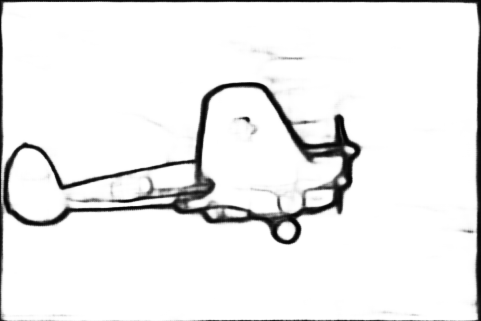}}
\end{subfigure}
\begin{subfigure}[b]{0.16\textwidth}
\centering
\frame{\includegraphics[width = \textwidth]{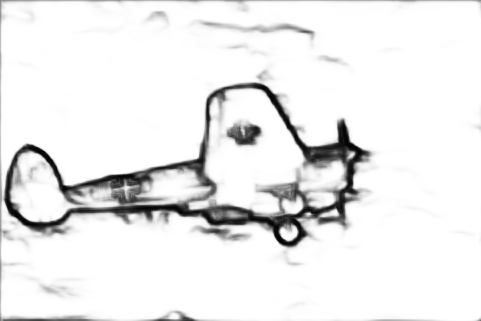}}
\end{subfigure}
\begin{subfigure}[b]{0.16\textwidth}
\centering
\frame{\includegraphics[width = \textwidth]{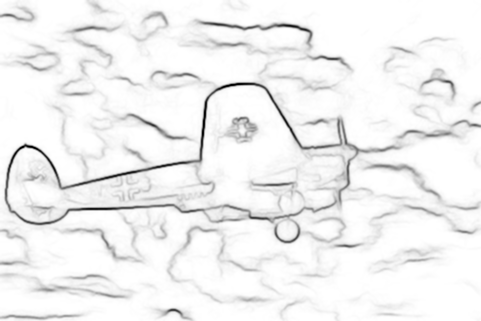}}
\end{subfigure}
\begin{subfigure}[b]{0.16\textwidth}
\centering
\frame{\includegraphics[width = \textwidth]{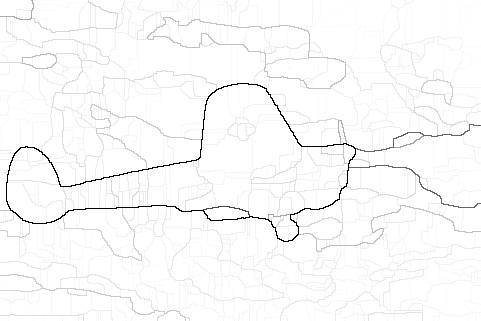}}
\end{subfigure}

\vspace{2pt}

\begin{subfigure}[b]{0.16\textwidth}
\centering
\includegraphics[width = \textwidth]{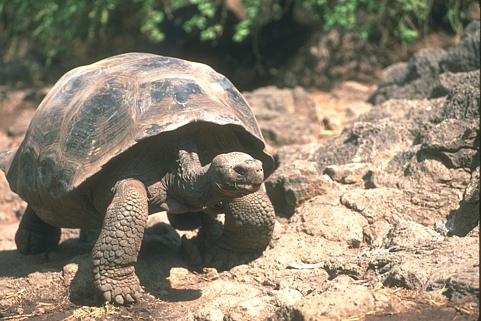}
\end{subfigure}
\begin{subfigure}[b]{0.16\textwidth}
\centering
\frame{\includegraphics[width = \textwidth]{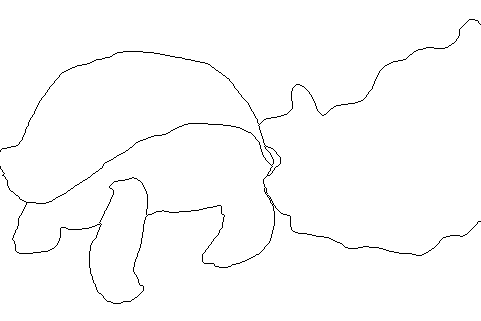}}
\end{subfigure}
\begin{subfigure}[b]{0.16\textwidth}
\centering
\frame{\includegraphics[width = \textwidth]{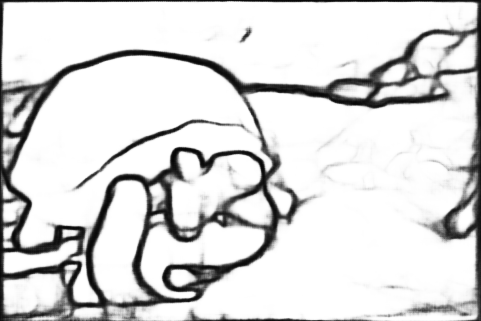}}
\end{subfigure}
\begin{subfigure}[b]{0.16\textwidth}
\centering
\frame{\includegraphics[width = \textwidth]{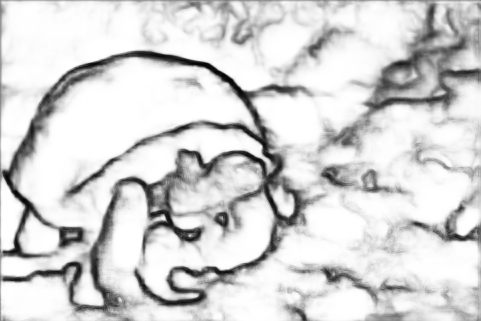}}
\end{subfigure}
\begin{subfigure}[b]{0.16\textwidth}
\centering
\frame{\includegraphics[width = \textwidth]{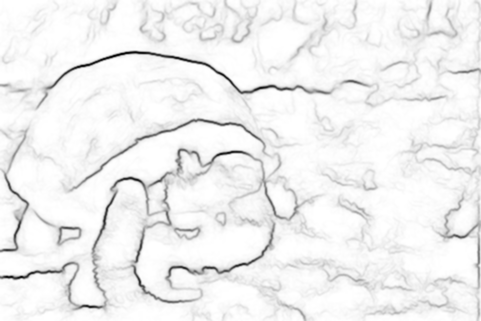}}
\end{subfigure}
\begin{subfigure}[b]{0.16\textwidth}
\centering
\frame{\includegraphics[width = \textwidth]{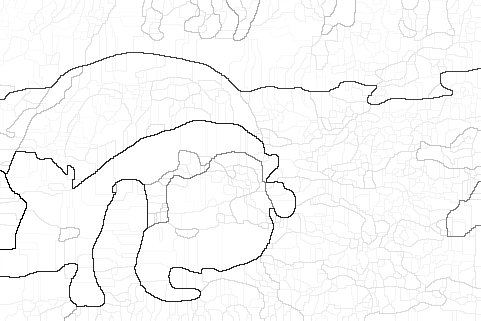}}
\end{subfigure}

\caption{Side-by-side comparison of edge detection algorithms. All edge images are originally returned by the algorithms before non-maximum suppression.}
\label{fig:visual-comparison}
\end{figure*}

\begin{figure*}
\centering
\begin{subfigure}[b]{0.03\textwidth}
\rotatebox{90}{\hspace{0.25cm} \textbf{Input Image}}
\end{subfigure}
\begin{subfigure}[b]{0.23\textwidth}
\centering
\frame{\includegraphics[width = \textwidth]{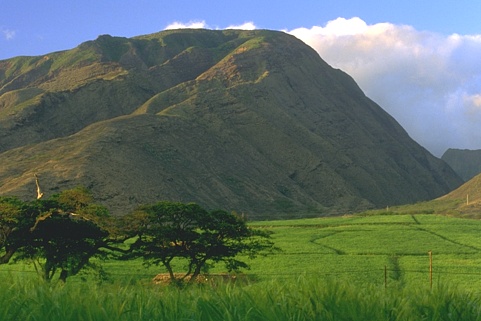}}
\end{subfigure}
\begin{subfigure}[b]{0.23\textwidth}
\centering
\frame{\includegraphics[width = \textwidth]{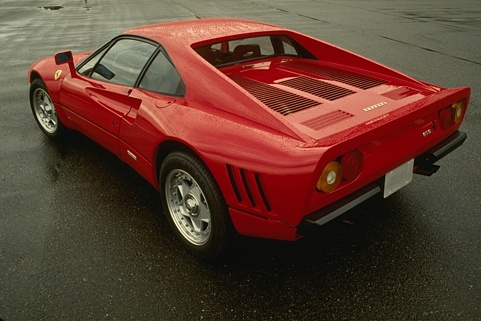}}
\end{subfigure}
\begin{subfigure}[b]{0.23\textwidth}
\centering
\frame{\includegraphics[width = \textwidth]{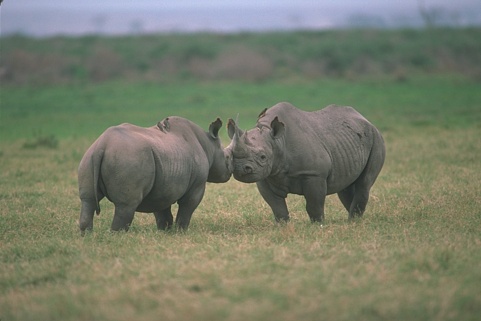}}
\end{subfigure}
\begin{subfigure}[b]{0.23\textwidth}
\centering
\frame{\includegraphics[width = \textwidth]{figures/BSDS500/images/103006.jpg}}
\end{subfigure}

\vspace{2pt}

\begin{subfigure}[b]{0.03\textwidth}
\rotatebox{90}{\hspace{0.80cm} $\mathbf{L=0}$}
\end{subfigure}
\begin{subfigure}[b]{0.23\textwidth}
\centering
\frame{\includegraphics[width = \textwidth]{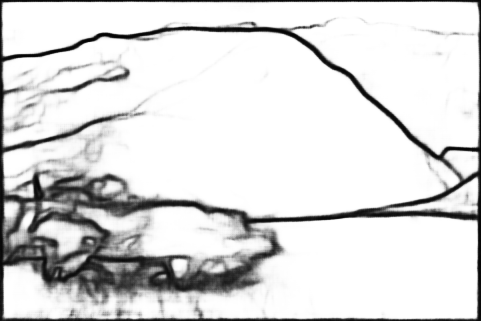}}
\end{subfigure}
\begin{subfigure}[b]{0.23\textwidth}
\centering
\frame{\includegraphics[width = \textwidth]{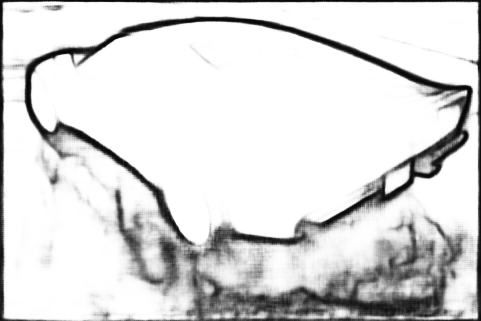}}
\end{subfigure}
\begin{subfigure}[b]{0.23\textwidth}
\centering
\frame{\includegraphics[width = \textwidth]{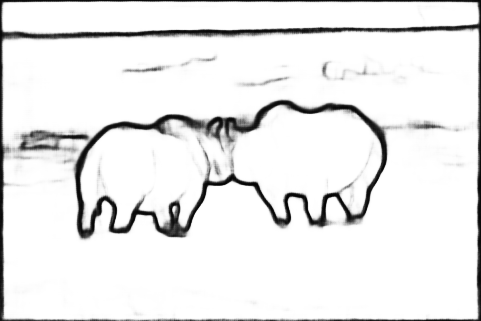}}
\end{subfigure}
\begin{subfigure}[b]{0.23\textwidth}
\centering
\frame{\includegraphics[width = \textwidth]{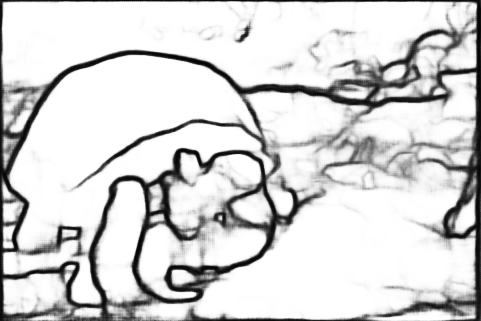}}
\end{subfigure}

\vspace{2pt}

\begin{subfigure}[b]{0.03\textwidth}
\rotatebox{90}{\hspace{0.80cm} $\mathbf{L=2}$}
\end{subfigure}
\begin{subfigure}[b]{0.23\textwidth}
\centering
\frame{\includegraphics[width = \textwidth]{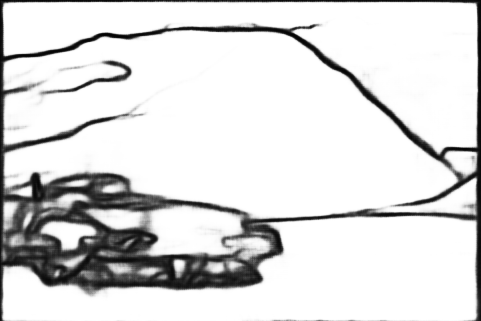}}
\end{subfigure}
\begin{subfigure}[b]{0.23\textwidth}
\centering
\frame{\includegraphics[width = \textwidth]{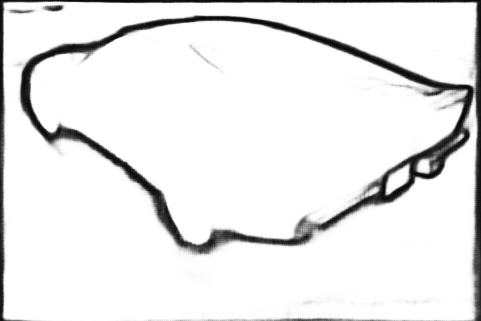}}
\end{subfigure}
\begin{subfigure}[b]{0.23\textwidth}
\centering
\frame{\includegraphics[width = \textwidth]{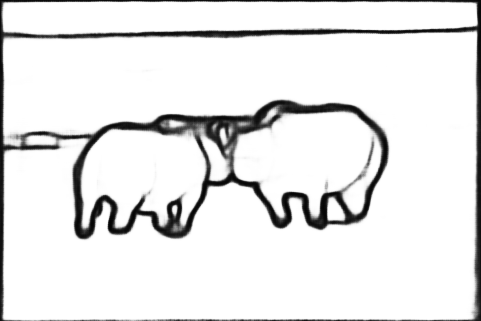}}
\end{subfigure}
\begin{subfigure}[b]{0.23\textwidth}
\centering
\frame{\includegraphics[width = \textwidth]{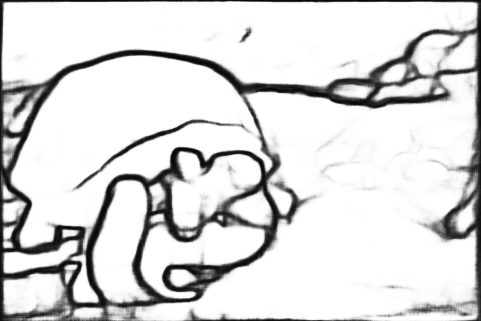}}
\end{subfigure}

\caption{The recursive network ($L=2$) improves the results of encoder-decoder network with skip-connections ($L=0$) by cleaning noisy edges and enhance stronger ones.}
\label{fig:recursive}
\end{figure*}

We evaluate our algorithm on BSDS500 \cite{Arbelaez2011}, NYUD \cite{Silberman2012} and Pascal Context \cite{Mottaghi2014} datasets using standard metrics such as ODS/OIS F-measure and AP. The BSDS500 dataset has edge annotation ground truth while the others do not. The NYUD and Pascal Context datasets are primarily for semantic segmentation. To obtain the ground truth edges, we first identify all the boundary pixels, treat them as a binary image and then apply image thinning using \texttt{MATLAB} function \texttt{bwmorph} (see examples in Fig. \ref{fig:how-to-get-ground-truth}).

\subsubsection{BSDS500}
The Berkeley Segmentation Dataset and Benchmark (BSDS500) \cite{Arbelaez2011} consists of $200$ training, $100$ validation and $200$ testing images. We use the training and validation sets ($300$ images) for training our RED-NET. Each colored image is of size $481 \times 321$ or $321 \times 481$ and is manually annotated ground truth contours. We simply overlay all annotations followed by image thinning to obtain a single ground truth image. Unlike other image-to-image deep learning framework such as HED \cite{Xie2015} which resizes the input image to a fixed size of $400 \times 400$, our RED-NET runs on original image without resizing. We use padding to make image dimension fit after convolutional, pooling and deconvolutional layers and crop the output to get the result of the original dimension.

\subsubsection{NYUD}
The NYUD dataset \cite{Silberman2012}, was used for edge detection in \cite{Ren2012,Gupta2014}, has $1449$ RGB-D images of indoor scenes (which are quite different from outdoor scenes of the BSDS500 \cite{Arbelaez2011}). As a result, it is more challenging because the edges are more cluttered and there are more variations. Here we use the setting described in \cite{Dollar2015} and evaluate our RED-NET on data processed by \cite{Gupta2014}. The NYUD dataset is split into $795$ training and $654$ testing images. These splits are carefully selected such that images from the same scene are only in one of these sets. All images are of size $640 \times 480$. This dataset also has depth image and although our RED-NET is easily extensible to RGB-D image, we do not use this information for our experiment. HED \cite{Xie2017} has three networks accepting RGB, depth encoded HHA \cite{Gupta2014} and RGB-HHA, respectively. Consequently, we include the results of all these three network versions in Table \ref{tbl:nyud}. For a fair comparison, during evaluation we increase the maximum tolerance allowed for correct matches of edge predictions to ground truth from $0.0075$ to $0.011$ as used in \cite{Gupta2014,Dollar2015,Xie2017}.

\subsubsection{Pascal Context}
The Pascal Context dataset \cite{Mottaghi2014} contains carefully localized pixel-wise semantic annotations for the entire image on the PASCAL VOC 2010 detection trainval set. It contains $\num{10103}$ images, which is approximately $20$ times larger than the BSDS500 dataset, span over $459$ semantic categories. Images in this dataset have various sizes and are quite challenging due to the increased scene complexities.

\subsection{Visual Comparison}
Fig. \ref{fig:visual-comparison} shows side-by-side comparison between different boundary detection algorithms. As we can see, the non-deep learning methods such as SE \cite{Dollar2015} and gPb-owt-ucm \cite{Arbelaez2011} produce sharp and clean edges in areas with high-contrast but fail at low-contrast regions because they only use local features and thus do not have a object-level understanding. 

The HED \cite{Xie2015}, which uses the features from VGG-16 network \cite{Simonyan2014}, performs much better and is able to capture objects even in low-contrast cases and it is not easily confused by object's interior boundary. Its weakness remains in the blurry and less localized edge responses, which may prevent it from recovering the sharp details.

Our RED-NET’s results are generally cleaner, sharper and more accurate 
Additionally, our results capture more global boundaries. For example, in the airplane image (second to last row of Fig. \ref{fig:visual-comparison}), only the most salient edges of the plane are retained.

Fig. \ref{fig:recursive} illustrates the benefits of the feedback loop in our network. In a pure encoder-decoder network without feedback loop (i.e. $L=0$), the results are blurry in the fine texture regions. However, with recursive network, these errors are cleaned up and salient edges are enhanced.

\subsection{Quantitative Comparison}
For numerical comparison, Table \ref{tbl:bsds500} shows the F-measure of various edge detection algorithms on BSDS500 dataset. It is obvious that our RED-NET is better than other methods regarding ODS/OIS F-measure with $\text{ODS} = 0.808$, $\text{OIS} = 0.828$ while providing reasonable $\text{AP} = 0.827$. Table \ref{tbl:nyud} provides the numerical statistics of the tested algorithms on the NYUD dataset. As we can see, although our RED-NET only takes as input RGB image and ignores depth information, it is still better than HED \cite{Xie2017}, RCF \cite{Liu2017} and COB \cite{Maninis2017} network which relies on both RGB and depth information. We set a new state-of-the-art edge detection on the NYUD dataset at $\text{ODS} = 0.793$, $\text{OIS} = 0.818$ and $\text{AP} = 0.832$.

The Pascal Context dataset is significantly larger and more challenging than the first two. From Table \ref{tbl:pascal-context}, without recursive network (i.e. $L=0$), our RED-NET will be similar to CEDN \cite{Yang2016}, except the skip-connections. As we can see from the table, skip-connections boost the ODS F-measure from $0.702$ to $0.744$, which is a huge improvement from CEDN even though it is still marginally behind COB \cite{Maninis2017}. However, with feedback loop (i.e. $L=2$), RED-NET edges out COB to achieve ODS F-measure of $0.759$. Furthermore, with our novel data augmentation of adding random Gaussian noise, RED-NET manages to push the results a little further at $\text{ODS} = 0.761$, $\text{OIS} = 0.785$ and $\text{AP} = 0.787$.

\subsection{Cross-dataset Evaluation}
To further demonstrate the generalization capability of our network, we train our model with one dataset and test it with another dataset. Table \ref{tbl:cross-dataset} shows the performance of our method on BSDS500, NYUD and Pascal Context datasets. The performance of a pretrained model is expected to be lower than that of a fine-tuned one. Our pretrained model yields a high precision but a low recall due to its object-selective nature between any two datasets. Furthermore, since BSDS500 dataset is pretty small and less diversified than the other two, model trained on it results in bigger drops in performance when tested on NYUD and Pascal Context.

\section{Conclusion}
We have proposed a method to substantially improve deep learning-based boundary detection performance. Our RED-NET adds skip-connections into the encoder-decoder network, which sharpens and preserves more details at the later layers, and a feedback loop, which allows progressive improvement of the edge image. We propose a novel data augmentation scheme use Gaussian blurring that can force the network to learn more salient edges as well as reduce the potential over-fitting. Our system is fully end-to-end trainable and operates in approximately $3.4$ frames per second, a speed of practical relevance. As measured on the standard datasets such as BSDS500, NYUD and Pascal Context, our RED-NET significantly outperforms other state-of-the-art approaches and sets new records in all three evaluation metrics. The source code and results will be made publicly available after the paper is accepted.

\section*{Acknowledgement}
We would like to thank the authors of BSDS500 dataset \cite{Arbelaez2011}, NYUD dataset \cite{Silberman2012} and Pascal Context dataset \cite{Mottaghi2014} for providing the benchmark data as well as evaluation toolbox. Additionally, we really appreciate the authors of all edge detection algorithms \cite{Canny1986,Felzenszwalb2004,Arbelaez2011,Lim2013,Ren2012,Dollar2015,Hallman2015,Kivinen2014,Ganin2015,Bertasius2015,Hwang2015,Shen2015,Xie2017,Kokkinos2016} for providing source code and/or benchmark results for comparison.

\bibliographystyle{IEEEtran}
\bibliography{references}

\end{document}